\documentclass[10pt,twocolumn,letterpaper]{article}

\usepackage[pagenumbers]{cvpr} 

\usepackage{graphicx}
\usepackage{amsmath}
\usepackage{amssymb}
\usepackage{booktabs}
\usepackage{graphicx}
\usepackage{wrapfig}
\usepackage{makecell}
\usepackage{xcolor}         
\usepackage{pifont}
\usepackage[font={small}]{caption}
\usepackage{mathtools}
\usepackage{mathtools, cuted}
\usepackage[misc]{ifsym}

\definecolor{antiquefuchsia}{rgb}{0.57, 0.36, 0.51}

\usepackage{amsmath}

\usepackage{multirow}
\usepackage{balance}
\usepackage[ruled,vlined]{algorithm2e}

\definecolor{citecolor}{RGB}{65,105,225}
\usepackage[pagebackref=true,breaklinks=true,letterpaper=true,colorlinks,citecolor=citecolor,bookmarks=false]{hyperref}

\usepackage[capitalize]{cleveref}
\crefname{section}{Sec.}{Secs.}
\crefname{section}{Section}{Sections}
\Crefname{table}{Table}{Tables}
\crefname{table}{Tab.}{Tabs.}

\begin{document}

\title{USV: Towards Understanding the User-generated Short-form Videos}
\author{Haoyue Cheng$^{1\dag}$
    \quad
    Su Xu$^{2\dag}$
    \quad
    Liwei Jin$^{1\dag}$
    \quad
     Wayne Wu$^{2}$
     \quad
    Chen Qian$^{2}$ 
    \quad
    Limin Wang\textsuperscript{1 \Letter}
    \quad\\
    $^{1}$State Key Laboratory for Novel Software Technology, Nanjing University, China \\
    \quad $^{2}$ SenseTime Research
    \\
    \quad
    {\tt\small \{xusu,wuwenyan,qianchen\}@sensetime.com}\\
    {\tt\small \{chenghaoyue98,liwei.jin97\}@gmail.com\quad lmwang@nju.edu.cn}
}

\maketitle

\begin{abstract}
Several large-scale video datasets have been published these years and have advanced the area of video understanding. However, the newly emerged user-generated short-form videos have rarely been studied. This paper presents USV, the User-generated Short-form Video dataset for high-level semantic video understanding. The dataset contains around 224K videos collected from UGC platforms by label queries without extra manual verification and trimming. Although video understanding has achieved plausible improvement these years, most works focus on instance-level recognition, which is not sufficient for learning the representation of the high-level semantic information of videos. Therefore, we further establish two tasks: topic recognition and video-text retrieval on USV. We propose two unified and effective baseline methods Multi-Modality Fusion Network (MMF-Net) and Video-Text Contrastive Learning (VTCL), to tackle the topic recognition task and video-text retrieval respectively, and carry out comprehensive benchmarks to facilitate future research. Our project page is \url{https://usvdataset.github.io}.
\footnote{$\dag$: Equal contribution,  \Letter: Corresponding author.}
\end{abstract}
\section{Introduction}

Recently, user-generated short-form videos from platforms such as TikTok~\cite{tiktok_official}, Kwai~\cite{kwai_official}, and Reels~\cite{reels_official} have drawn much attention~\cite{tiktok}. Understanding user-generated short-form videos is of great importance for practical usages such as video recommendation~\cite{davidson2010youtube}, venue analysis~\cite{youtube_revenue} and automated video summarization~\cite{ji2019query}. Take the video recommendation as an example. Zhu \etal~\cite{zhu2013videotopic} recommend videos by recalling videos that most fit the interested topic distribution of users. Deng \etal~\cite{deng2015twitter} take a real-time hot topic detection as the first step of recommendation.
However, neither a dataset nor a benchmark exists in previous literature for understanding user-generated short-form videos from a multi-modality and high-level semantic perspective.



In these years, many video datasets have been proposed and pushed the boundary of video understanding. For example, some~\cite{kay2017kinetics,kuehne2011hmdb,liu2017pku,monfort2019moments,oh2011large,shao2020finegym, soomro2012ucf101} are built for action recognition, others~\cite{gu2018ava, Heilbron2015ActivityNetAL,jiang2014thumos} are built for action localization. They mainly focus on recognizing instance-level actions and entities in long-term videos. None of the aforementioned works collect data purely from user-generated short-form videos, nor for leveraging multi-modality cues to facilitate understanding of high-level semantic information of videos.

User-generated short-form videos have four main features that distinguish them from other video forms. 
1) \textit{Topic Concentration}: 
User-generated short-form videos are more topic concentrated~\cite{nie2019multimodal} compared to professional-generated ones such as movies and TV series. Because short-form video platforms often have a duration constraint, short-form videos are forced to convey a single main topic in a few seconds.
2) \textit{Text Richness}:
User-generated short-form videos usually include much text information, such as titles, subtitles, dialogue, and comments. These texts are all user-generated and have rich semantic information related to the videos.   
3) \textit{High Activity}: 
User-generated video platforms are highly active, with millions of videos uploaded every day, together with all kinds of new topics of videos. This difference makes manually filtering the noise for a clean dataset trivial compared with scaling up with the noise, which makes the traditional time-consuming data process with manual annotating impractical.
4) \textit{Large Diversity}: 
User-generated short-form videos have more unique genres (\eg, slides, lectures, podcasts) and narratives (\eg, selfies, portraits) to demonstrate their topics.
With the large diversity, the multi-modality cues are rather important for understanding.

In this study, we aim to move towards understanding the user-generated short-form videos based on the aforementioned features. First, we introduce a new dataset named User-generated Short-form Video (USV-1.0). Specifically, USV-1.0 (the first version of USV) contains around 224K videos of 212 topic categories. Considering the first feature of \textit{Topic Concentration}, we propose a new task named topic recognition. Regarding the second feature of \textit{Text Richness}, we define the second task as video-text retrieval. Detailed definition and motivation of these two tasks can be found in Sec.~\ref{sec:4.1}. For the third feature of \textit{High Activity}, we use no human labor for verification and trimming but directly assign the queried words as the topics and the user-generated titles as the text to retrieve. This web-supervised scheme is good at leveraging the large-scale user-generated video stream to facilitate video understanding in the real world. The fourth feature of \textit{Large Diversity} identifies the main challenge of our tasks: how to integrate sufficient cues from various modalities to boost high-level semantic video understanding?


For topic recognition, we propose a simple yet effective baseline method Multi-Modality Fusion Network (MMF-Net) as a baseline. Specifically, MMF-Net is a three-branch network that fuses models' predictions for three modalities to form a consensus on the topic. For video-text retrieval, we adopt a video-text contrastive learning (VTCL) framework, which has proven effective in self-supervised learning. For both tasks, we build a comprehensive benchmark to facilitate future research. Benchmark experiments are conducted with our implementation based on a common protocol and evaluated under a unified setting without bells and whistles.

To be brief, our contributions are three-fold:
\begin{enumerate}
    \item New Data: we collect a new dataset: USV-1.0, the first large-scale dataset that aims to push the boundary of real-world short-form video understanding to our knowledge.
    \item New Tasks: we define a new task called topic recognition, and we first try to perform video-text retrieval based on user-generated titles. Both tasks focus on understanding high-level semantic information.
    \item New Methods: we propose MMF-Net and VTCL, the first trials to utilize both audio and subtitles to tackle the topic recognition task and the video-text retrieval respectively, and build a comprehensive benchmark to facilitate future research.
\end{enumerate}
\section{Related Work}
\noindent\textbf{Video Recognition Datasets.} Several video datasets have been published these years. From initial trials~\cite{kuehne2011hmdb,soomro2012ucf101} to large-scale benchmarks ~\cite{kay2017kinetics,monfort2019moments,monfort2019multi}. Then datasets for specific fields emerged, such as fine-grained gym datasets~\cite{ibrahim2016hierarchical,shao2020finegym}, human gestures~\cite{materzynska2019jester}, surveillance footage~\cite{oh2011large} and RGB-D camera~\cite{shahroudy2016ntu}. Also, datasets of a tremendous scale~\cite{abu2016youtube,diba2019holistic,karpathy2014large} with the help of automatic annotation systems and web data have been derived. Most of these listed restrict their label space to be instance-level visual entities, and most videos are generated by professionals.

A similar idea of topical understanding is observed in YouTube8M~\cite{abu2016youtube}. However, they also restrict their topic entities to be visual and collected from YouTube with mostly long-term videos that are hard to infer a single topic or retrieve by a single title. This difficulty forces them to label a video with several topics and eventually degrades into visual-instance recognition.

A few works have been done towards understanding short-form videos~\cite{liu2017towards,nie2019multimodal,nie2017enhancing,wei2019neural,zhang2020low}. However, most of their works are based on Vine, a video platform that has been shut down for years. Besides, the task designed on those datasets is scene/venue classification, which is still an instance-level recognition task.

Ours is not restricted to visual entities and is labeled video-wisely and purely from user-generated content.


\noindent\textbf{Video-Text Retrieval.} Learning videos with language has been a trend towards understanding videos, and video-text retrieval is one of the fundamental tasks. Common datasets including~\cite{rohrbach2017movie,xu2016msr,zhou2018towards} are relatively small, and~\cite{miech2019howto100m}~is too large to leverage and restricted to tutorial videos. Many of them focus on specific domains, such as instructional videos, cooking videos, \etc. Besides, most of them are well annotated and trimmed but not scalable.

As for methods, latent space-based models are common~\cite{kiros2014unifying,lin2014visual,miech2019howto100m,torabi2016learning}. Visual and textual representations are projected into a shared embedding space, where similarity can be measured directly. The typical visual encoding approach first extracts frame-level features and then aggregates them into video-level representation~\cite{torabi2016learning,lei2021less,torabi2016learning}. 
A similar paradigm for textual encoding is to extract each word feature and aggregate them into sentence feature~\cite{devlin2018bert,miech2018learning,otani2016learning,yu2018joint,yu2017end}. 
We aggregate frame-level visual features and encode sentence features directly to balance performance and computation cost.

\noindent\textbf{Self-Supervised Video Representation Learning.}
Self-supervised representation learning constructs different kinds of supervision tasks from the data itself, to learn semantic representation to promote downstream tasks. For video, some of these tasks include temporal ordering of videos~\cite{fernando2017self,lee2017unsupervised}, predicting motion and appearance~\cite{wang2019self}, predicting the other parts of videos~\cite{han2019video}, \etc. 

Contrastive learning has been widely used in self-supervised representation learning, which aims at distinguishing similar and dissimilar data pairs.
For example, Radford~\etal~\cite{radford2021learning}~learns visual and textual embedding based on image-text pair-wise contrastive learning, breaking through the limitation of classifying only on predefined categories in traditional image classification.
Nowadays, more and more works leverage multi-modality data based on contrastive learning~\cite{arandjelovic2017look,korbar2018cooperative,miech2020end,miech2019howto100m,sun2019learning}. Korbar~\etal~\cite{korbar2018cooperative}~uses visual and audio correspondence in semantic and temporal dimensions to construct positive and negative samples. Some methods make use of the abundant source of text, such as tags, labels, ASR, scripts, \etc. Miech~\etal~\cite{miech2020end}~differs from previous approaches in employing multiple positive samples, denoted as MIL-NCE. Inspired by previous works, we conduct contrastive learning on video-text retrieval task. 

\section{USV: User-Generated Short-form Video Dataset}
Our project aims to build a dataset for user-generated short-form video understanding, which is both intractable in task and data itself.
We will first demonstrate the procedure of building the dataset in Sec.~\ref{sec:3.1} and illustrate the challenges within the dataset. Afterward, we will give statistics and comparison with other datasets in Sec.~\ref{sec:3.2}.

\subsection{How the Dataset is Built}
\label{sec:3.1}
USV-1.0 is built by first pre-defining the taxonomy and using it as the query words to collect videos. We then extract the visual, audio, and textual modalities from the raw videos. We label the videos in an untrimmed and unverified manner that performs no extra manual annotating. 

\noindent\textbf{Stage I: Categories Taxonomy.}
To our best knowledge, no literature studies the semantic taxonomy for user-generated short-form videos. Most datasets~\cite{abu2016youtube,Heilbron2015ActivityNetAL,kay2017kinetics} built their taxonomy regarding some former sociological researches and picked the visual-dependent ones. However, we do not adopt this as it will neglect the important feature of videos: \textit{Large Diversity}. In addition, words from the knowledge graph or other references are outdated. One important feature of user-generated short-form videos is that they are \textit{Highly Active} with trending topics coming up daily such as \textit{ASMR (Autonomous sensory meridian response)}, \textit{block-chain}, \textit{finger dance}, which can't be found in any knowledge graph of a sociological study on the taxonomy of videos. Therefore, we refer to the sector system of several online video platforms and
pick 32 macro topics as the root nodes of our taxonomy, including \textit{anime}, \textit{international affairs}, \textit{sport}, \textit{affection}, \etc.

\begin{figure}[t]

\begin{center}
   \includegraphics[width=\linewidth]{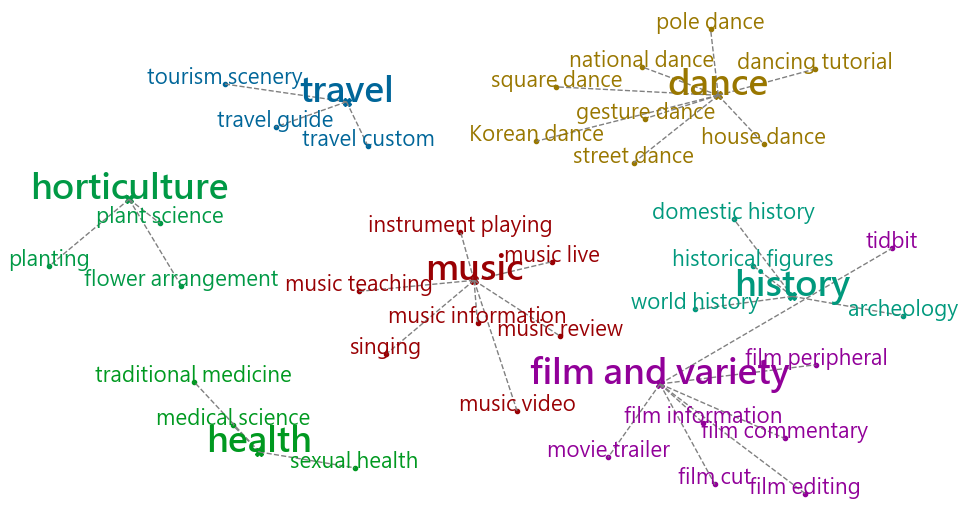}
\end{center}
\vspace{-0.4cm}
   \caption{\textbf{The word embedding t-SNE of the taxonomy.} We select a part of the taxonomy for a better presentation. Different colors represent different macro-categories. Macro-categories are largely distant, while intra-category distance is short.}
\label{fig:tsne}
\end{figure}

To step further, we investigate the top-watched categories of each macro topic in UGC short-form video platforms, and expand each into several micro topics as leaf nodes. As a result, we obtain the final 212 leaf nodes. Note that topics are not limited to visual-only ones, and they can be an abstract concept (\eg, \textit{affection}), audio- (\eg, \textit{ASMR}) or textual- (\eg, \textit{international news}) dependent, which requires understanding from multi-modality aspects. An overview of the topic taxonomy is demonstrated by t-SNE~\cite{maaten2008visualizing} in Fig~\ref{fig:tsne}.

\noindent\textbf{Stage II: Collecting.}
\label{stage:II}
We collect the videos by the 212 micro-categories. We use the words of the 212 micro topics and their synonyms as the queries to retrieve videos and their corresponding titles, and assign the queried topics as the labels of the videos. The rationale behind it is the feature \textit{Topic Concentration} as the query word tends to be the only topic of the video. We are aware of the noise that emerges from this query-based 
collecting and labeling strategy. The candidate videos are recalled by the internal recommendation service of UGC platforms. Therefore some videos may be hardly related to the queried topics. Therefore, our dataset is noisy and challenging. Then the unique video id is used to skip duplication within each micro topic. We collect $231226$ videos in total.

\noindent\textbf{Stage III: Modality Extraction.}
We extract three modalities: visual, audio, and text from all raw videos. We choose the raw soundtrack and the subtitles on the raw frames as the representation of audio and text modality. We extract the audio with FFMPEG~\cite{ffmpeg}; as for text, we sample one frame each second and perform OCR detection with EasyOCR~\cite{easyOCR} to extract subtitles. Trivial words such as the water-print are filtered, and then we approximate the subtitles of the sampled frames to their adjacent frames.



\noindent\textbf{Stage IV: Human Verification.}
Human verification can be a challenge for video tasks. First, the total duration shown in Tab.~\ref{tab:dataset_comparison} of our dataset is \textgreater 100d for a human to go through, which will take more than years. Second, crowded source annotators have personal bias and recognition differences, especially when asked to perform high-level inference rather than simply identify instances. Third, as stated in the introduction about the \textit{High Activity} feature, the incremental ability of the data possessing strategy is far more critical than the correctness of supervision signals for highly active media like short-form UGC videos. We only verify the validation and test set for topic labels by two human annotators. The annotation user interface asks the annotators: By watching/listening/reading the video, whether the main topic of the video is the same as the query word. If both annotators choose \textit{NO}, the video is considered label noise and removed from the validation set. The original validation set of size $31226$ is randomly split from the total dataset. After verification, $6776$ videos are excluded. Since the validation set has the same distribution as the training set, we can also assume that there are approximately 21.7\% noisy labels in the training set. Similar verification is also applied to the test set. After that, We remove the validation and test sets of videos with empty titles for the video-text retrieval task.

\begin{table}\small
\caption{\textbf{Statistics summary of USV-1.0 dataset.} We have at least 216 training videos for each category and an average number of 1058.}
\vspace{-0.4cm}
\label{tab:dataset_summary}
\begin{center}
\begin{tabular}{l c}
\toprule
\multicolumn{2}{c}{\textbf{Dataset Specifications}}\\ %
\midrule
Number of videos & 224,450\\
Train set & 200,000\\
Validation set & 24,450\\
\midrule
Number of micro-category labels & 212\\
Average number of videos per micro-category & 1,058\\
Number of macro-category labels & 32\\
Average number of videos per macro-category & 7,014\\
Range of training videos per micro-category & 216-1,786\\
\midrule
Average length of subtitles & 106\\
Average length of titles & 32\\
\midrule
Average duration of videos(in seconds) & 55\\
Total length(in days) & 143 \\
\bottomrule
\end{tabular}
\end{center}
\end{table}

\begin{figure}
\vspace{-0.2cm}
\begin{center}
   \includegraphics[width=0.9\linewidth,height=1.0\linewidth]{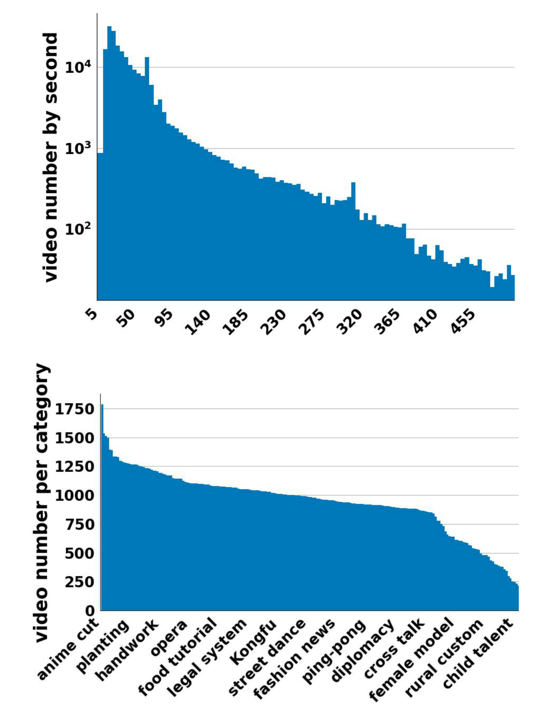}
\end{center}
\vspace{-0.4cm}
\caption{\textbf{Video number and duration distribution.} \textbf{Top:} distribution of the number of videos for each duration. \textbf{Bottom:} number of videos for each category.}
\label{fig:merged}
\end{figure}

\subsection{Datasets Comparison}
\label{sec:3.2}
USV-1.0 is a large-scale dataset with various categories and contains rich modality data. More details are listed in Tab.~\ref{tab:dataset_summary}~and Fig.~\ref{fig:merged}. We compare ours with other intensively studied video recognition datasets~\cite{abu2016youtube,diba2019holistic,goyal2017something,Heilbron2015ActivityNetAL,karpathy2014large,kay2017kinetics,kuehne2011hmdb,monfort2019moments,ray2018scenes,soomro2012ucf101} and video-text retrieval datasets~\cite{anne2017localizing,damen2018scaling,krishna2017dense,miech2019howto100m,rohrbach2017movie,xu2016msr,zhou2018towards} in Tab.~\ref{tab:dataset_comparison} and ~\ref{tab:dataset_comparison_retrieval}.
We demonstrate the characteristics of our dataset: non-visual-only, topical, and user-generated.

\begin{figure}[t]

\begin{center}
   \includegraphics[width=0.9\linewidth]{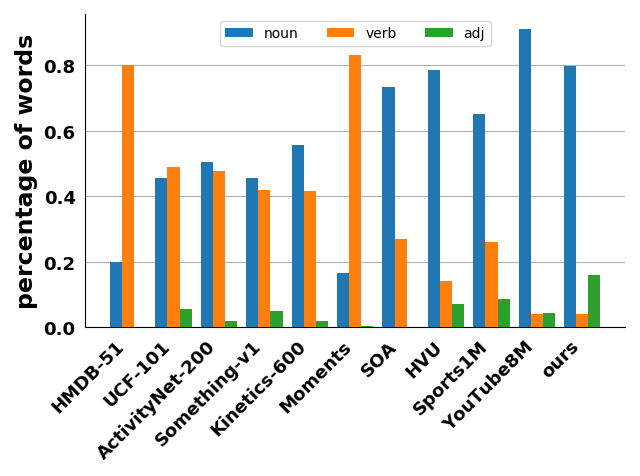}
\end{center}
\vspace{-0.4cm}
\caption{\textbf{Comparison of part-of-speech.} Our dataset taxonomy mainly consists of nouns and adjectives because of high-level semantics, so does YouTube8M. We use an NLP tool named textblob~\cite{textblob} to determine the part of speech for each word.}
\label{fig:part-of-speech}

\end{figure}
\noindent
\textbf{Non-visual-only.}
In Tab.~\ref{tab:dataset_comparison}, \textbf{V} (Visual-only) represents whether the categorization can be done by visual modality only, and inversely, $\neg{\textbf{V}}$ is non-visual-only. Note that although some datasets such as Kinetics and YouTube8M preserve the audio soundtrack, they are also visual-only because the videos or classes, depending on other modalities for classification, are removed by human annotators. UCF-101 is not visual-only since it preserves audio-dependent classes and samples like \textit{playing instruments}. Distinctly, categorization on the USV-1.0 dataset largely relies on multi-modality data since the data itself is \textit{diverse}, thus our dataset is $\neg{\textbf{V}}$.

\noindent
\textbf{Topical.}
\textbf{T} (Topical) denotes whether the dataset taxonomy focuses on grabbing the topic or recognizing the existing instances in videos. As illustrated in Fig.~\ref{fig:part-of-speech}, our label space has a different distribution compared with most other datasets. Former datasets focus on human actions and interactions between humans and objects. Thus their label spaces are composed mostly of nouns and verbs. On the contrary, since our taxonomy is constructed by topics that summarize the videos' main ideas, they usually consist of attributive adjuncts and the key objects of the video. Therefore, nouns and adjectives have much larger proportions. YouTube8M has a similar distribution to ours as it is also topical. Although nouns account for a large proportion of HVU and SOA\footnote{Since SOA can't be accessed publicly, we consider \textit{scenes} and \textit{objects} as nouns and \textit{actions} as verbs.}, these labels are all instance-wise.


\noindent\textbf{User-generated.}
\textbf{U} (User-generated) indicates whether all videos come purely from UGC video platforms. User-generated short-form videos have several unique characteristics that require to be studied respectively. USV-1.0 is constructed purely from the UGC platforms. In contrast, other datasets are collected from platforms like YouTube where the majority of videos are released by professional video producers. EPIC-KITCHENS~\cite{damen2020epic} is recorded by 32 individuals via headsets but not from a platform with a large number of users, so we consider it not user-generated.


\begin{table}\large

\caption{\textbf{Dataset comparison(topic recognition).} We compare the total number of videos and clips, the number of categories, total duration in time, whether the video categorization depends on other modalities besides vision($\neg{\textbf{V}}$), whether the label taxonomy is Topical(\textbf{T}), and whether videos totally come from User-generated video platform(\textbf{U}).}
\vspace{-0.4cm}
\label{tab:dataset_comparison}
\begin{center}
\resizebox{\columnwidth}{!}{
\begin{tabular}{l c c c c c c c}
\toprule
\textbf{Dataset} & \textbf{Videos} &
\textbf{Clips} &
\textbf{Categories} & \textbf{Duration} &  \textbf{$\neg$V} & \textbf{T} & \textbf{U}\\
\midrule
HMDB-51~\cite{kuehne2011hmdb} & 3.3k & 6.7k & 51 & 5.7h & $\times$ & $\times$ & $\times$\\
UCF-101~\cite{soomro2012ucf101} & 2.5k & 13k & 101 & 27h & \checkmark & $\times$ & $\times$\\
ActivityNet-200~\cite{Heilbron2015ActivityNetAL} & 20k & 28k & 200 & 27d & $\times$ & $\times$ & $\times$\\
Something(v1)~\cite{goyal2017something} & 108k & 108k & 174 & 121h & $\times$ & $\times$ & $\times$\\
Kinetics-600~\cite{carreira2018short} & 495k & 495k & 600 & 57d & $\times$ & $\times$ & $\times$\\
Moments~\cite{monfort2019moments} & 1M & 1M & 339 & 31d & \checkmark & $\times$ & $\times$\\
SOA~\cite{ray2018scenes} & 562k & 562k & 65d & 553 & $\times$ & $\times$ & $\times$\\
HVU~\cite{diba2020large} & 577k & 577k & 4378 & 66d & $\times$ & $\times$ &$\times$\\
Sports1M~\cite{KarpathyCVPR14} & 1M & 1M & 487 & 10y & $\times$ & $\times$  & $\times$\\
YouTube8M~\cite{abu2016youtube} & 8M & 8M & 4800 & 57y & $\times$ & \checkmark & $\times$\\
\midrule
USV-1.0 (\textbf{Ours}) & 224K & 224k & 212 & 143d & \checkmark & \checkmark & \checkmark\\
\bottomrule
\end{tabular}}
\end{center}
\end{table}

\begin{table}[!]\large

\caption{\textbf{Dataset comparison(video-text retrieval).} We compare several common video-text retrieval datasets with ours in the number of videos/clips, average number of sentences per video/clip, total dataset duration, average duration per video, whether the videos are User-generated(\textbf{U}).}
\vspace{-0.4cm}
\label{tab:dataset_comparison_retrieval}
\begin{center}
\resizebox{\columnwidth}{!}{
\begin{tabular}{l c c c c c}
\toprule
\textbf{Dataset} & \textbf{Videos} & \textbf{Clips} & \textbf{Captions} & \textbf{Duration} & \textbf{U}\\
\midrule
MSR-VTT(v1)~\cite{xu2016msr} & 7k & 10k & 200k & 40h & $\times$\\
LSMDC~\cite{rohrbach2015dataset} & 200 & 128k & 128k & 150h & $\times$\\
YouCook2~\cite{zhou2018towards} & 2k & 14k & 14k & 176h & $\times$\\
EPIC-KITCHENS~\cite{damen2020epic} & 432 & 40k & 40k & 55h & $\times$\\
DiDeMo~\cite{anne2017localizing} & 10k & 27k & 41k & 87h & $\times$\\
ANet-Captions~\cite{krishna2017dense} & 20k & 100k & 100k & 849h & $\times$\\
HowTo100M~\cite{miech2019howto100m} & 1.2M & 136M & 136M & 15.3y & $\times$\\
\midrule
USV-1.0 (\textbf{Ours}) & 224k & 224k & 224k & 143d & \checkmark\\
\bottomrule
\end{tabular}}
\end{center}
\vspace{-0.2cm}
\end{table}

\section{User-Generated Short-Form Video Understanding}
To benchmark user-generated short-form video understanding on USV-1.0 dataset, we propose two specific tasks called topic recognition and video-text retrieval. We will first demonstrate the detailed definition and motivation of these two tasks in Sec.~\ref{sec:4.1}. Then we describe our proposed \textit{Multi-Modality Fusion Network (MMF-Net)} for the topic recognition task in Sec.~\ref{sec:4.2}. Afterward, we provide a simple but effective \textit{Video-Text Contrastive Learning (VTCL)} framework for video-text retrieval task in Sec.~\ref{sec:4.3}. The methods we proposed are relatively simple but sufficient. The reason is that we intend to conduct preliminary explorations to provide  insights on how multi-modality cues are beneficial for holistic user-generated short-form video understanding. 

\subsection{Task Definition and Motivation}
\label{sec:4.1}
\noindent
\textbf{Topic Recognition Task.} Although both topic recognition and action recognition~\cite{Heilbron2015ActivityNetAL,kay2017kinetics,kuehne2011hmdb,monfort2019moments,soomro2012ucf101} can be categorized as a single-label multi-class classification problem, there are two crucial points to distinguish them. First, topic recognition uses topics as labels, which contain more high-level semantic information than most instance-level classification problems. Second, our proposed topic recognition encourages using multi-modality information inside videos for classification. To be specific, raw frames, audios, and subtitles can be used during the training and evaluating stages. Modality-based tools like ASR or OCR are also not forbidden. Thus, topic recognition is not a purely instance-level visual task, but a multi-modal high-level semantic video classification task.

\noindent
\textbf{Video-Text Retrieval Task.} Most user-generated short-form videos are paired with user-uploaded titles, which are usually strongly related to the corresponding videos. We view these collected titles as natural weak video captions. These ``captions" are not annotated by professional annotators and are easy to scale. Moreover, personal bias may be relieved with a large variety of ``annotators". Formally, our defined title-based video-text retrieval consists of two subtasks: text-based video retrieval and video-based text retrieval. Suppose a test set of $n$ pairs of videos and titles, text-based video retrieval aims to find the corresponding video for every given title in this set, and video-based text retrieval is vice versa. Similar to topic recognition, video-text retrieval encourages utilizing multi-modality information too. Our proposed video-text retrieval task forms a harder problem than topic recognition and its well annotated counterpart, since titles are usually of large diversity and sometimes related to videos at a high semantic level, e.g., a video of a beach traveling VLog with a title \textit{``Happy holiday"}, in which case the title and visual frames are only related in high semantic level. To summarize, video-text (user-generated title) retrieval calls for high-level semantic video understanding rather than instance-level recognition. 

\noindent

\noindent\textbf{Tasks Design Motivation.} Our intention for building USV is to push the boundary of high-level semantic UGC short video understanding. The reason why we choose topic recognition rather than existed tasks such as video object detection, action recognition or video classification is that, those tasks probe the ability of learning low-level video representation, while we figure it is more demanded by the industry to develop a sophisticated model to be able to reason along modalities and time to get a overall representation of videos.

Video-text retrieval steps further to some extent, as it abandons predefined labels in the classical supervised learning scheme and uses natural language as supervisory signals. Video-text retrieval is not the only way to leverage natural language information to help video understanding, and generating tasks like video captions or text-based video generation can also bind video with natural language. So why is retrieval? Here is an intuitive explanation: we notice that babies can learn a concept by matching pictures and texts, but it is hard for them to write sentences or draw pictures. So it is natural to assume video-text retrieval is a moderately difficult task for current video understanding.

\subsection{Topic Recognition}
\label{sec:4.2}
\begin{figure}[t]
\centering
\includegraphics[width=0.9\linewidth]{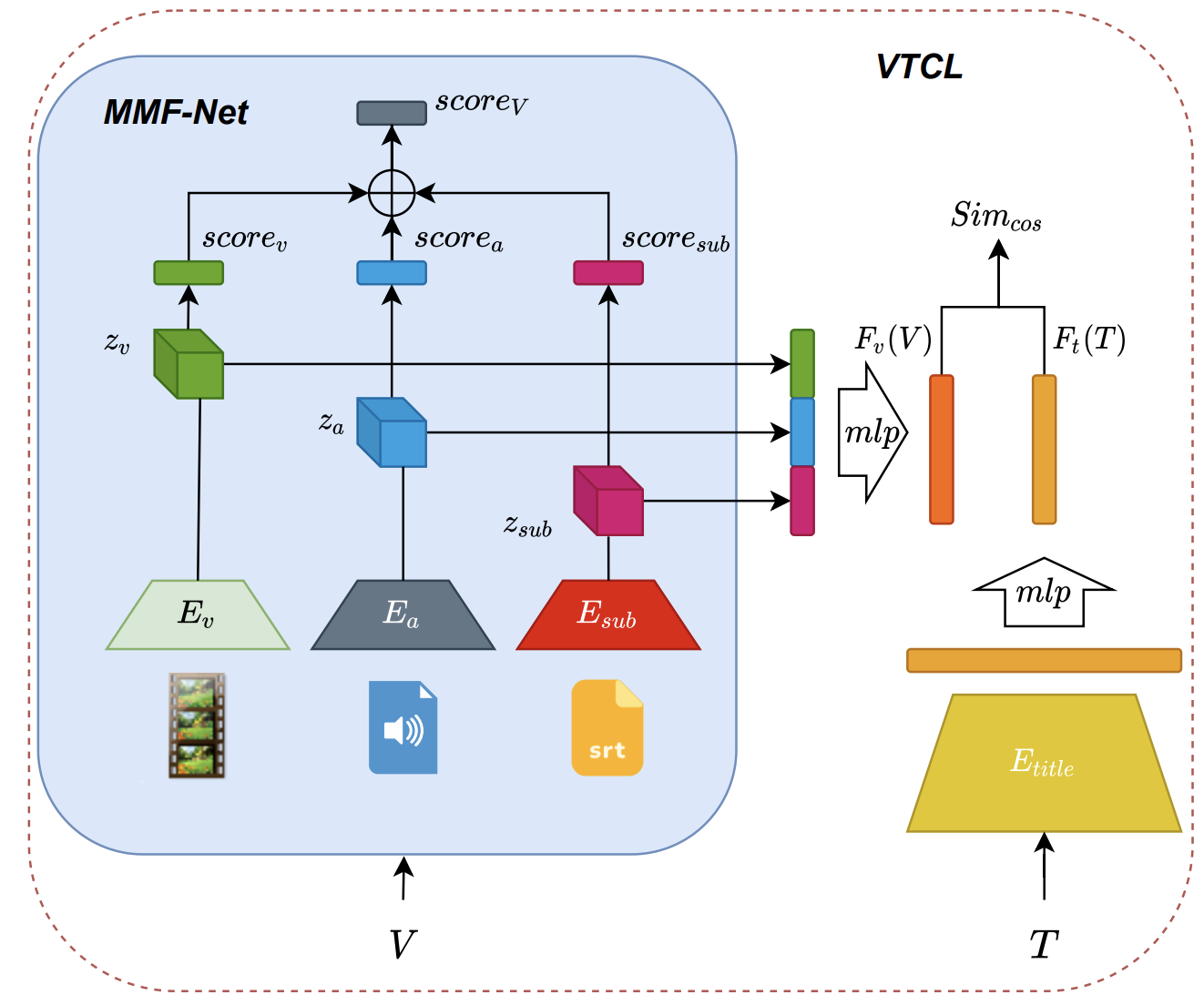}
\caption{\textbf{The pipeline of our Multi-Modality  Fusion  Network(MMF-Net) and video-text contrastive learning (VTCL) framework for topic recognition and video-text retrieval. } First, the multi-modality signals are fed into modality-specific networks for feature extraction. For topic recognition, these features are used to predict 212-d classification scores separately, and these scores are 
fused to form a video-level prediction. For video-text retrieval, multi-modality features are concatenated and projected by an MLP layer into a v-t joint-embedding, in which video features are matched with text features of user-generated titles via cosine similarity.}
\label{fig:pipeline}
\end{figure}

We regard topic recognition as a fully-supervised learning problem and solve it with a general but effective network architecture named \textit{Multi-Modality Fusion Network (MMF-Net)}. As is shown in Fig.~\ref{fig:pipeline}, MMF-Net can be abstracted as a three-stream late fusion network to combine the results of three multi-modality streams. Late fusion has proven to be a simple but powerful technique used by many video recognition networks~\cite{feichtenhofer2019slowfast,simonyan2014two}. We detail the specific structures of these three branches in the following.


\noindent
\textbf{Branch I: Visual Encoder.} Visual branch is implemented with classical 2D- and 3D-Conv networks such as TSN~\cite{wang2016temporal}, I3D~\cite{carreira2017quo}. It consists of a feature extractor (backbone) built up with 2D- or 3D-Conv modules and a classifier (head) built with a linear layer. Those models have demonstrated good representation learning ability on action recognition datasets~\cite{goyal2017something, kay2017kinetics}, which demands models to utilize both spatial and temporal information.

\noindent
\textbf{Branch II: Audio Encoder.} Following the work of~\cite{arandjelovic2017look,xiao2020audiovisual}, we use a 2D image backbone to extract the audio features, namely ResNet-18 and log-Mel spectrograms of clips as input to generate the prediction logits. Log-Mel spectrograms have been used for audio recognition~\cite{xu2018large}, multi-modality self-supervised learning~\cite{arandjelovic2017look,shukla2020learning}, \etc, and have proven to be a discriminative and compact representation of audio. What's more, using spectrograms can largely benefit from the excellent designs of 2D image backbones like ResNet~\cite{He_2016_CVPR} from numerous image research areas.

\noindent
\textbf{Branch III: Text Encoder.} Many new genres of videos have derived from short-form video platforms. For example, one popular type is those informative lecture videos produced by semi-professional self-media producers. We observe that most user-generated short-form videos are associated with subtitles, which can be viewed as an inner part of videos and are essential to holistic video understanding. Video subtitles are extracted by a powerful and robust open-source OCR toolkit named EasyOCR~\cite{easyOCR}. We use EasyOCR to extract frame-level subtitles and concatenate these subtitles in chronological order to form a video-level subtitle classification dataset. The textual branch consists of a pre-trained multilingual BERT model~\cite{devlin2018bert} appended with a linear classification head. We fine-tune this textual branch on the subtitle classification dataset.  

\subsection{Video-Text Retrieval}
\label{sec:4.3}

As topic labels are not available under the setting of video-text retrieval, we are only given $n$ pairs of videos and titles, forming a scenario of cross-modality self-supervised learning.

\noindent
\textbf{Video-Text Contrastive Learning Framework.} Following recent visual-language self-supervised learning methods~\cite{li2020learning,miech2020end,radford2021learning}, we propose an end-to-end contrastive learning framework called \textit{Video-Text Contrastive Learning (VTCL)}. As is shown in Fig.~\ref{fig:pipeline}, we aim to learn two mapping functions $F_v$ and $F_t$ that map videos and titles into a $d$-dimensional joint embedding space. In this $d$-dimensional joint embedding space, the embedding of a certain video is pulled closer to the embedding of its corresponding title, and the distance between embeddings of unrelated videos and titles should be extended. We measure the distance between video and title embeddings by cosine similarity:
\begin{equation}
    s(V, T)=\frac{\langle \mathbf{F_v}(V), \mathbf{F_t}(T)\rangle}{\|\mathbf{F_v}(V))\|_{2}\|\mathbf{F_t}(T)\|_{2}}
\end{equation}
where $V$ stands for sampled clips of a certain video and $T$ stands for a certain title.
Following ~\cite{miech2020end,radford2021learning}, we adopt an InfoNCE loss widely used in recent contrastive learning works:
\begin{equation}
    \mathcal{L}_i = - \log \frac{e^{s(V_i,T_i)/\mathcal{T}}}{\sum_{j=1}^{n}(e^{s(V_i,T_j)/\mathcal{T}}+e^{s(V_j,T_i)/\mathcal{T}})}
\end{equation}
where $\mathcal{T}$ is a temperature parameter, and $n$ is the size of the mini-batch.

\noindent
\textbf{Video Encoder.} The three branches of the video encoder share the same design as described in Sec.~\ref{sec:4.2}. There are two different characteristics worth mentioning: 1. The visual branch only uses 8-frame TSN~\cite{wang2016temporal} for its high performance in topic recognition and computing cost balance; 2. The parameters of the language model are frozen when training. Because the corpus for BERT pre-training is much larger and cleaner than ours, fine-tuning BERT directly on USV might cause language model crushing. The features output by three branches are concatenated and then projected by an MLP composed of 2 linear layers into d-dimensional joint embedding space.

\noindent
\textbf{Text Encoder.} The text encoder also utilizes the above-mentioned multilingual BERT to extract 768-dimension features from user-generated titles. The BERT is also frozen during training for the same reason. These features are then projected by an MLP composed of 2 linear layers into d-dimensional joint embedding space, too.
\section{Experiments}
We design fundamental and important experiments for two tasks on USV-1.0, which are illustrated as follows. Besides, we have explored more different experiment settings and case studies in the supplementary materials.
\subsection{Experiment Setup} For supervised topic recognition, our experiment setup is unified for a fair comparison. Most of the training and testing follow the protocols in the original papers unless specified. We use 8 RGB frames of scale 224 for training, either sparsely or densely sampled. For evaluation, an identical amount of total frames as input is used, and mean class accuracy (mca) is adopted as the metric since our validation set is label unbalanced. We weight the class scores of vision, audio, and text branches according to empirical weight parameters of 1, 0.5, and 0.5 to obtain the ensemble classification scores, since vision plays a more important role in recognition than audio and text.

For self-supervised video-text retrieval, we use the same video input and the training setting as topic recognition unless specified. The temperature parameter $\mathcal{T}$ is set as 0.05. To evaluate our learned VTCL embedding, we randomly select 20k valid video-title pairs from the validation set and divide them into 20 subsets evenly. Following the evaluating setting of previous retrieval datasets~\cite{rohrbach2017movie,xu2016msr,zhou2018towards}, we average the standard recall metrics R@1, R@5, R@10, and the median rank(Median R) on these 20 subsets.



\subsection{Topic Recognition}

\begin{table}[!h]
\caption{\textbf{Baseline Performance.} The column \textit{Input frames} is formed by ($num\_crops \times clip\_length \times frame\_interval \times num\_clips$). $-f3$ denotes the frames used for training is 3.}
\label{tab:baseline}
\vspace{-0.4cm}
\begin{center}
\resizebox{1.0\columnwidth}{!}{
\begin{tabular}{l l c c c}
\toprule
\textbf{Method} & \textbf{Type} & \textbf{pre-trained}& \textbf{Input frames}& \textbf{mca}\\
\midrule
TSN-f3 & 2D & Scratch &10x1x1x24 & 70.13\\ 
TSN-f5 & 2D & Scratch&10x1x1x24 & 71.15\\ 
TSN-f8 & 2D & Scratch&10x1x1x24 & \textbf{73.51} \\ 
TSN-f8 & 2D & ImageNet&10x1x1x24 &  71.75\\ 
TSN-f8 & 2D & Kinetics-400&10x1x1x24 & 71.73\\ 
I3D & 3D & Scratch&3x8x8x10 & 66.85 \\ 
R(2+1)D & 3D & Scratch&3x8x8x10 & 63.52 \\ 
SlowFast & 3D & Scratch&3x32x2x10 & 70.00 \\ 
\bottomrule
\end{tabular}}
\end{center}
\end{table}
\noindent\textbf{Baseline Models.}
For baseline experiments, we choose 5 mostly used models for video recognition, which are TSN~\cite{wang2016temporal}, TSM~\cite{lin2019tsm}, I3D~\cite{carreira2017quo}, R(2+1)D~\cite{tran2018closer}, Slowfast~\cite{feichtenhofer2019slowfast}. All of them are using ResNet-50 as the base model. 3D-Conv models are trained with 8 densely sampled frames as input, while 2D models are trained with 8 sparsely sampled frames. The evaluation inputs are controlled to be the same (except for SlowFast which has double streams with different amounts of input) for a fair comparison. It is usually expected that 3D models may have better performance based on experience on several large-scale datasets~\cite{kay2017kinetics}. However, according to Tab.~\ref{tab:baseline}, it is not the case on USV-1.0. The 3D models are surpassed by the 2D models with a great margin: The best-performed model is TSN trained with 8 frames, while the best 3D model is SlowFast which requires much more input frames that still scores around 3.5\% lower than TSN.

\noindent\textbf{Training Frames.}
In addition, we also evaluate the result of TSN trained with different input frames, from 3 to 8. A steady increment can be found, which indicates that for a better understanding of the main topic of USV-1.0, more frames in training can be beneficial.

\noindent\textbf{Pre-training.}
Furthermore, we first pre-train the best performed TSN model on ImageNet and Kinetics-400 and fine-tune it on USV-1.0 until convergence. It turns out the pre-trained weights even have a negative impact on USV-1.0. The performance of ImageNet pre-trained and Kinetics-400 pre-trained scores have a marginal difference and are lower than the from-scratch one by approximately 2\%. This finding suggests that there may exist a considerable domain gap between ours and Kinetics-400 and ImageNet.

\noindent\textbf{MMF-Net.} In Tab.~\ref{tab:MMF-Net}, we demonstrate the effectiveness of the design of MMF-Net. For Branch I, the visual recognition branch, we have evaluated two typical video models of different mechanisms. We observe that multi-modality branches have a positive impact. To be more specific, audio and text branches score lower than 50\%, however they bring no overall harm but benefit when fused with the visual branch. Note that the coefficients of all branches are set heuristically rather than by finding the optimized combination. 

\begin{table}[h]\small
\caption{\textbf{MMF-Net performance.} We evaluate the effect of each branch when fused to the visual branch as an ablation study.}
\vspace{-0.4cm}
\label{tab:MMF-Net}
\begin{center}
\begin{tabular}{l l c}
\toprule
\textbf{Branch} & \textbf{Modality} & \textbf{mca($\Delta$)}\\
\midrule
I(TSN) & V& 73.51\\ 
I(Slowfast) & V& 70.00\\ 
\midrule
II & A & 40.88\\ 
I(TSN) + II  & V + A & 74.71(+1.20)\\
I(Slowfast) + II  & V + A & 71.81(+1.81)\\
\midrule
 III & T & 46.61\\ 
I(TSN) + III  & V + T & 78.18(+4.67)\\ 
I(Slowfast) + III & V + T & 76.07(+6.07)\\ 
\midrule
 I(TSN) +  II +  III  & V + A + T & \textbf{78.84(+5.33)}\\
 I(Slowfast) +  II +  III  & V + A + T & \textbf{77.11(+7.11)}\\ 
\bottomrule
\end{tabular}
\end{center}
\end{table}

\subsection{Video-Text Retrieval}
\label{sec:5.3}
We adopt different multi-modality settings progressively by adding audio and subtitle streams to the video branch separately and together. All these variations use the same training setting for a fair comparison.      

Similar to topic recognition, we find multi-modality information also has positive effects on VTCL. As shown in Tab.~\ref{tab:retrieval}, additional audio and subtitle information fused with the visual branch respectively outperform the visual-only baseline consistently. Moreover, combining the three modalities results in the highest performance among all the variations. It shows that integrating multi-modality information may help understand user-generated short-form videos from a holistic perspective.

\noindent
\textbf{Retrieval-Based Zero-Shot Topic Recognition.} We evaluate our multi-modality embedding of VTCL with zero-shot classification on USV without any fine-tuning in Tab.~\ref{tab:zero_shot}. We transform class labels and videos into the same embedding space and recognize the video as the class with the highest cosine similarity. Although retrieval-based zero-shot topic recognition performs worst than its fully-supervised counterpart, it is proposed as a generic self-supervised learning strategy using only user-generated titles to conduct topic recognition. This scheme is easy to scale up and generalize to open-world applications.
\begin{table}\large
\caption{\textbf{Video-text retrieval performance.} We evaluate the impact of using different combinations of three modalities in the video encoder of our proposed VTCL as an ablation study. In order to distinguish, we denote subtitles as T and titles as Title, respectively.}
\label{tab:retrieval}
\vspace{-0.4cm}
\begin{center}
\resizebox{\columnwidth}{!}{
\begin{tabular}{l c c c c}
\toprule
\textbf{Modality} & \textbf{Recall@1} & \textbf{Recall@5} & \textbf{Recall@10} & \textbf{Median R} \\
\midrule
V to Title & 20.29 & 46.20 & 56.49 & 5.00  \\
V + A to Title & 22.19 & 48.58 & 59.08 & 4.00  \\
V + T to Title & 21.98 & 49.18 & 60.28 & 4.00  \\
V + A + T to Title & \textbf{23.51} & \textbf{50.77} & \textbf{62.20} & \textbf{3.00}  \\
\midrule
Title to V & 20.77 & 46.37 & 56.83 & 5.00  \\
Title to V + A  & 22.33 & 48.90 & 59.85 & 4.00  \\
Title to V + T  & 22.45 & 49.24 & 60.18 & 4.00  \\
Title to V + A + T & \textbf{23.71} & \textbf{51.12} & \textbf{62.30} & \textbf{3.00}  \\
\bottomrule
\end{tabular}}
\end{center}
\end{table}

\begin{table}\small
\caption{\textbf{Retrieval-based zero-shot topic recognition performance.} We evaluate the impact of using different combinations of three modalities in the video encoder as an ablation study.}
\vspace{-0.4cm}
\label{tab:zero_shot}

\begin{center}
\begin{tabular}{lc}
\toprule
\textbf{Modality} & \textbf{mca($\Delta$)}\\
\midrule
V & 27.21 \\
V + A & 27.23(+0.02) \\
V + T & 27.23(+0.02) \\
V + A + T & \textbf{29.01(+1.80)}\\
\bottomrule
\end{tabular}
\end{center}
\end{table}

\section{Conclusion}
In this work, we build the first large-scale user-generated short-form video dataset, define topic recognition and video-text retrieval tasks, and propose the MMF-Net and VTCL framework as simple but effective baselines for these two tasks. We conduct comprehensive experiments as preliminary explorations to facilitate future research on user-generated short-form video understanding.

\appendix

\section*{Appendix}

\section{Implementation Details}
\label{sec:1}
\subsection{MMF-Net Formalization}
To formalize MMF-Net, we define our dataset as $\mathcal{V} = \left\{ v_{i} \right\}$, where $v_{i}$ denotes the i-th video in the dataset. Similarly, we denote ${x}_{i}^{j(k)}$ as the j-th clip-wise input sampled with a fixed duration from a total of $N$ clips of the i-th video's k-th modality. For the subtitle branch, we use video-level text input namely $N=1$. For example, $N=5$,  $i=10$,  $j=3$, $k=2$, $x_{i}^{j(k)}$ represents the 3-rd audio clip out of 5 uniformly divided clips of the 10-th video. With all annotations above, the inference procedure of \textit{MMF-Net} can be formalized as:
\begin{equation} \label{eq:1}
    \hat{s}_{i}^{(k)} =
    \frac{1}{N} * \sum_{j=1}^{N} \sigma(E^{(k)}(x_{i}^{j(k)}))
\end{equation}
\begin{equation}\label{eq:2}
    \hat{{S_{i}}} = \frac{1}{3} *  \sum_{k=1}^{3} w_{k} \hat{s}_{i}^{(k)}
\end{equation}
where $\hat{s}_{i}^{(k)}$ denotes the recognition score of the k-th branch on the i-th video, $E^{(k)}$ denotes the feature extractor of branch $k$, and $w_k$ denotes the weight of the score of branch $k$. Specifically, we use 1, 0.5, 0.5 as the weights from Branch I to III. Eq. (\ref{eq:1}) can be viewed as a clip-wise consensus~\cite{wang2016temporal} with the average operation, and Eq. (\ref{eq:2}) is a branch-wise late-fusion for video topic recognition by averaging as well.

While training, we set $N=1$ for each video and we train each branch separately, therefore the forward equation and loss function for each branch during training is:
\begin{align}
    \hat{s}_{i}^{(k)} &= \sigma(E^{(k)}(x_{i}^{(k)}))\\
    \mathcal{L}_\text{rec} &= - \sum_{c=0}^{211} y_{c} * \ln \hat{s}_{c}
\end{align}
$y_{c}$ denotes the ground truth for the c-th logits from 0 or 1. The loss is a plain Cross-Entropy loss with $Softmax$ activation.
\subsection{Branch Instantiation Details}
\paragraph{Branch I}
For 2D backbones such as TSN, we divide each video into 8 parts and randomly select one frame from each part for training. Frames are resized to short-side 256p first, and perform a multi-scale crop into a square, and resized to 224 $\times$ 224; While testing, we sample 24 frames uniformly, each performing a short-size resizing to 256 and a ten-crop to a square 256p image. TSN is instantiated by the backbone of ResNet-50 and a linear layer as the classification head mapping the global average pooled 2048-d vector to 212 class scores; TSM follows the original optimal setting is the original paper~\cite{lin2019tsm}, with $1/8$ channels shifted on each $conv1$ layer of all child blocks in each ResNet stage.

For 3D backbones such as Slowfast, we randomly sample a clip of 8 frames of interval 8 and perform the identical augmentation as 2D backbones; For testing, 10 uniform clips are sampled, and each performs a three-crop to 256p frames. Clip-wise predictions are then averaged to form a video-wise prediction. I3D is based on a ResNet-50 backbone and inflates the 2D-Convs into 3D ones. Note that we do not bootstrap 3D filters from 2D pretrains but train them from scratch; For R(2+1)D, it follows the instantiation of I3D, but only factorizes all 3D-Convs into (2+1)D ones; As for SlowFast, both branches are a ResNet3D backbone such as I3D. The input for the slow pathway is $8\times 8$, while for the fast path it is $32\times 2$, and the base channels of the fast are $1/8$ of the slow one. The 2048-d and 512-d feature vectors of the two pathways are concatenated and mapped by a linear layer to the 212-d score.

\paragraph{Branch II}
For each video, we randomly sample a corresponding clip of 2s, \ie $2\times16000$ bins. Then the log-Mel transformation with 80 mel-filters, window size 32ms, hop size 16ms and $fft$ size of 1280 will turn the 2-second clip into a spectrogram of 80 channels and 128 time stamp, \ie a gray image of size $1\times 80 \times 128$ ; When testing, we uniformly sample 10 clips per video and average all clips' logits to get the final prediction.

\paragraph{Branch III}
We use the pre-trained multilingual BERT model provided by Google-Research~\cite{devlin2018bert} with 104 languages, 12-layer, 768-hidden, 12-heads, and 110M parameters as the text-branch backbone. For each video, we use EasyOCR~\cite{easyOCR} to extract subtitles of one frame per second and filter out watermarks. Then we concatenate frame-level OCR results in chronological order to form video-level subtitles as the raw input. There are 12.3k videos with valid subtitles out of 20k training videos. We just leave them empty strings as BERT input. For topic recognition, we fine-tune the pre-trained multilingual BERT model with an initial learning rate is 5e-5 for 8 samples per batch; a linear scheduler with 500 warm-up steps is applied until saturation. For video-text retrieval, we freeze the pre-trained multilingual BERT model as an off-shelf feature extractor when training and evaluating.

\subsection{Experiment Setup}
\noindent\textbf{Training Settings.}
For topic recognition, 2D-Conv models are trained for 100 epochs until saturation, the initial learning rate of 0.05 and a step schedule that decay the learning weight by 0.1 at step 40 and 80, the batch size is 128; 3d-Conv models are trained for 200 epochs, the initial learning rate is 0.1 for 64 samples per batch. The scheduler used is cosine annealing until saturation. This setting follows several public codebases including ~\cite{mmaction2,fan2020pyslowfast}, in which 2D models use fewer epochs as the sparse sampling can lead to faster convergence, and smaller and plainer learning since 2D models have fewer parameters.

For video-text retrieval, TSN is applied for the visual branch I in the video encoder. Frozen multilingual BERT is used as the feature extractors both in subtitle branch III and the title encoder. VTCL is trained for 100 epochs, with an initial learning rate of 0.03, a cosine annealing scheduler, and a batch size of 128. VTCL is under-optimized for computing cost limitations.

\noindent\textbf{Evaluation Settings and Metrics.}
As examined by multiple pieces of research ~\cite{carreira2017quo,fan2020pyslowfast,wang2016temporal}, the number of input frames greatly impacts accuracy. Therefore, we balance the number of crops and clips to control an identical number of frames as input for all experiments.

Since our validation set is of the same distribution as the training set, namely the numbers of videos of each topic category are not identical, it is unfair to use the top k accuracy. Therefore, we used the mean class accuracy for evaluation on USV and top-1 accuracy on downstream datasets.

To evaluate VTCL on video-text retrieval, we randomly select 20k valid video-title pairs from the validation set and divide them into 20 subsets evenly. Following the evaluating setting of previous retrieval datasets~\cite{rohrbach2017movie,xu2016msr,zhou2018towards}, we average the standard recall metrics R@1, R@5, R@10, and the median rank(Median R) on these 20 subsets. 

\section{Supplementary Analysis of Experiments}
\label{sec:2}
\subsection{Confusion Analysis}
In Tab.~\ref{tab:TSN confusions}, \ref{tab:Slowfast confusions} and~\ref{tab:MMFNet confusions}, we demonstrate the confusions of the worst ten classes for three models: TSN~, SlowFast, and MMF-Net (with TSN as Branch I), which are the representative methods for 2D-Conv, 3D-Conv, and ours. The confusion shows that fine-grained categories are easier to be confused, such as \textit{planting} and \textit{farm work}, \textit{pet cat}, and \textit{pet dog}. This observation raises the challenge of accurate spatial discrimination; Classes such as \textit{restaurant reviews} and \textit{food reviews} may have a very similar visual, audio, and textual appearance, and it further emphasizes the importance of the reasoning ability of the model.
\begin{table}[htbp]\scriptsize
\begin{center}
\resizebox{0.48\textwidth}{!}{
\begin{tabular}{c c c }
\toprule
\textbf{Class 1} & \textbf{Class 2} & \textbf{Confusion}\\
\midrule
male model & layman handsome influencer & 65\% \\
restaurant review & food review & 49\% \\
planting & farm work & 44\% \\
movie information & movie review & 31\% \\
global military intelligence & domestic military intelligence & 29\% \\
science anecdote & cutting edge of science \& technology & 26\% \\
rural performance & folklore & 26\% \\
luxury car & roadster & 24\% \\
roadster & luxury car & 21\% \\
military exercise & global military intelligence & 21\% \\

\bottomrule
\end{tabular}
}
\end{center}
\vspace{-0.2cm}
\caption{\textbf{Top-10 class confusions in USV-1.0, using the TSN model.}}
\label{tab:TSN confusions}
\end{table}
\begin{table}[htbp]\scriptsize
\begin{center}
\begin{tabular}{c c c }
\toprule
\textbf{Class 1} & \textbf{Class 2} & \textbf{Confusion}\\
\midrule
restaurant review & food review & 45\% \\
planting & farm work & 44\% \\
movie information & movie review & 34\% \\
male model & layman handsome influencer & 34\% \\
roadster & luxury car & 31\% \\
rural performance & folklore & 26\% \\
domestic military intelligence & global military intelligence & 24\% \\
pet cat & pet dog & 22\% \\
layman beauty influencer & hairdressing & 22\% \\
financial management & stock market & 21\% \\
\bottomrule
\end{tabular}
\end{center}
\vspace{-0.2cm}
\caption{\textbf{Top-10 class confusions in USV-1.0, using the Slowfast model.}}
\label{tab:Slowfast confusions}
\end{table}

\begin{table}[htbp]\scriptsize
\begin{center}
\resizebox{0.48\textwidth}{!}{
\begin{tabular}{c c c }
\toprule
\textbf{Class 1} & \textbf{Class 2} & \textbf{Confusion}\\
\midrule
male model & layman handsome influencer & 93\% \\
restaurant review & food review & 49\% \\
movie information & movie review & 41\% \\
planting & farm work & 41\% \\
luxury car & roadster & 35\% \\
science anecdote & cutting edge of science \& technology & 35\% \\
parent-child interaction & children's sport & 34\% \\
domestic military intelligence & global military intelligence & 24\% \\
layman handsome influencer & layman beauty influencer & 20\% \\
global military intelligence & domestic military intelligence & 20\% \\

\bottomrule
\end{tabular}
}
\end{center}
\vspace{-0.2cm}
\caption{\textbf{Top-10 class confusions in USV-1.0, using the MMF-Net model.}}
\label{tab:MMFNet confusions}
\end{table}
\subsection{Quantitative Effect of Multi-modality}

\begin{figure*}[htbp]
\begin{center}
  \includegraphics[width=0.9\linewidth]{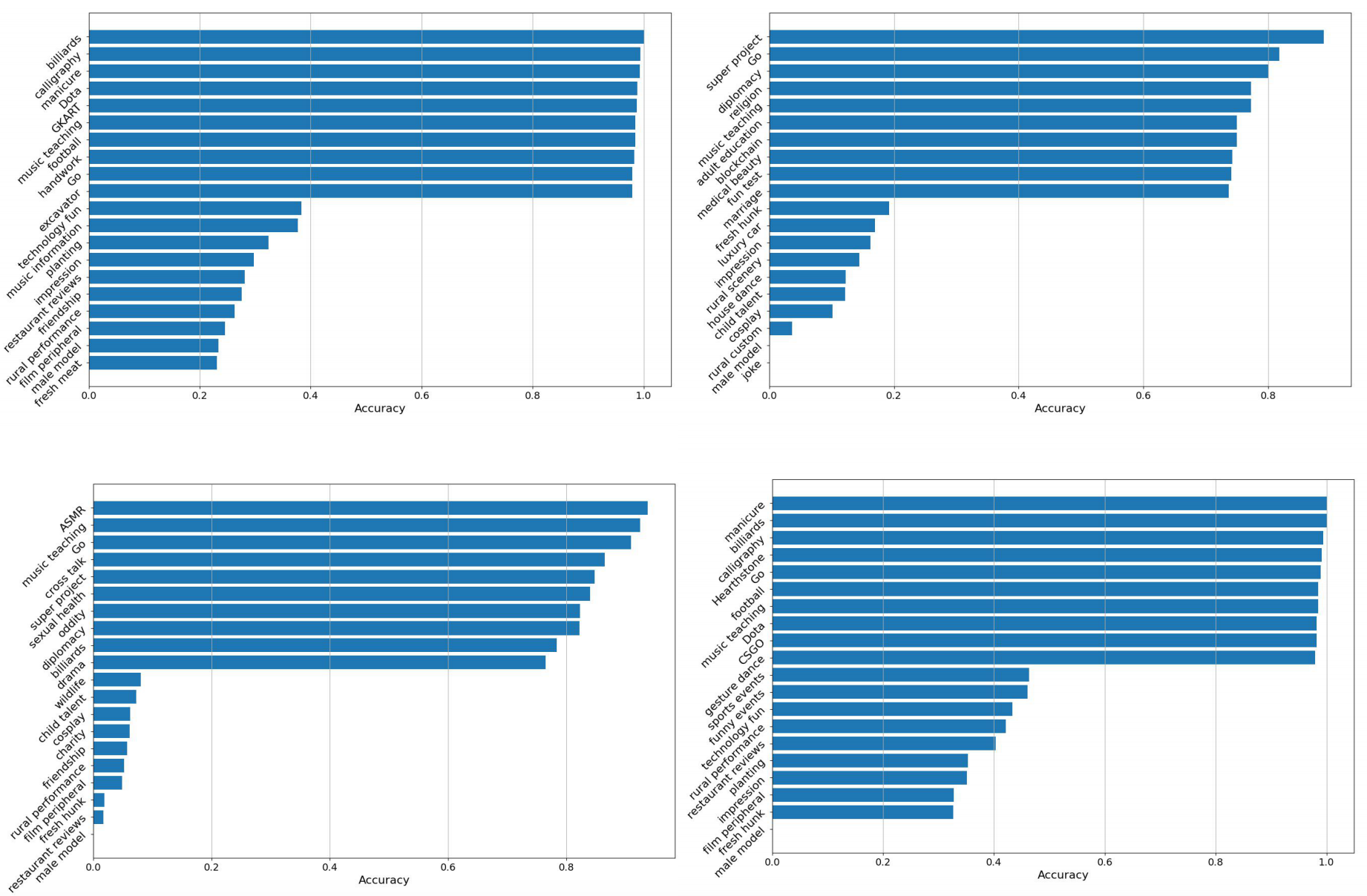}
\end{center}
\vspace{-0.2cm}
  \caption{\textbf{Top-10 easy and hard classes.} \textbf{Upleft}: Visual branch. \textbf{Upright}: Textual Branch.
 \textbf{Downleft}: Audio Branch.
 \textbf{Downright}: Fused.
  }
\label{fig:easy_and_hard}
\end{figure*}
We demonstrate the quantitative effectiveness of multi-modality design by showing the easiest and hardest 10 classes for each multi-modality branch and their fusion. It can be observed in Fig.~\ref{fig:easy_and_hard} that classes with special visual clues such as \textit{billiards} with similar green tables, \textit{Dota} with similar game interface rank the best when using a single visual branch only; classes that have special audio features rank high for the audio branch. For example, \textit{ASMR} which is a newly emerged videos in which authors make soft sounds with an extreme close sound field, and classes such as \textit{music teaching} and \textit{cross-talk} also rank higher than other branches; For the text branch, informative videos that usually possess detailed subtitles such as \textit{diplomacy}, \textit{adult education}, \textit{blockchain} are also among the bests.

We further study the gain and loss caused by each modality in Tab.~\ref{tab:modality_change}. We show 10 classes of the greatest accuracy gain and drop for Branch I when fused with Branch II, III, and both. 
Besides the gain for those acoustic and informative classes mentioned before, we observe that the textual branch brings a dominant gain for the whole, indicating the importance of visual-textual learning in UGC short-form videos.

\begin{table}[t]\tiny
\begin{center}
\resizebox{0.48\textwidth}{!}{
\begin{tabular}{l c c c }
\toprule
\textbf{Model} & \textbf{+Audio}($\Delta mca$) & \textbf{+Text}($\Delta mca$) & \textbf{+Both}($\Delta mca$) \\
\midrule
\multirow{10}*{TSN}& movie mashup 0.15& blockchain 0.34 & blockchain 0.39\\
                  &music video 0.14 & religion 0.27 & adult education 0.27\\
                  &impression show 0.13 & adult education 0.25& hunting 0.24\\
                  &ACG dance 0.13 & friendship 0.22 & friendship 0.22\\
                  &tourist guide 0.10& hunting 0.22 & sociality 0.20\\
                  \cline{2-4}
                  &male model -0.21 & male model -0.17 & male model -0.23\\
                  &luxury car -0.18 & grange -0.05 & domestic military intelligence -0.11\\
                  &primary and secondary -0.12 & luxury car -0.05&luxury car -0.09 \\
                  &archaeology -0.09 & domestic military intelligence -0.05 & swim -0.04\\
                  &grange -0.07 & child talent -0.04 &tourism attraction -0.04 \\

\midrule
\multirow{10}*{SlowFast}
                  &movie review 0.12 &religion 0.40 & adult education 0.39\\
                  &rural custom 0.11 & adult education 0.35& blockchain 0.34\\
                  &ASMR 0.10 & blockchain 0.28& religion 0.31\\
                  &singing 0.09 & friendship 0.25& college 0.27\\
                  &K-pop dance 0.09 & college 0.25& sociality 0.25\\
                  \cline{2-4}
                  &male model -0.19 &male model -0.14 & male model -0.27\\
                  &luxury car -0.11 & live music -0.03& luxury car -0.07\\
                  &restaurant review -0.08 & spoof -0.03& sports star -0.03\\
                  &roadster -0.07 &handsome influencer -0.01& spoof -0.03\\
                  &hunting -0.06 &sports event -0.01  & swim -0.02\\

\bottomrule
\end{tabular}
}
\end{center}
\vspace{-0.2cm}
\caption{\textbf{Accuracy gain and loss.} Each column denotes the top-5 classes with the greatest gain and loss due to one or both modalities are fused.}
\label{tab:modality_change}
\end{table}

\section{Detailed Taxonomy}
\label{sec:3}
We list the complete taxonomy of topic categories here. Words in bold are among the 32 macro topics, while those listed below are the micro topics expanded from the macro topic.

{
\noindent
\textbf{1. entertainment} \\
\indent    entertainment scene\\
\indent	entertainment gossip\\
\textbf{2. variety show}\\
\indent   variety show\\
\textbf{3. film and television}\\
\indent    movie clip\\
\indent	movie mashup\\
\indent	titbits\\
\indent	movie information\\
\indent	movie review\\
\indent	movie peripheral derivatives\\
\indent	movie trailor\\
\textbf{4. amusing}\\
\indent    spoof\\
\indent	weirdo\\
\indent	joke\\
\indent	roast\\
\indent	funny dubbing\\
\indent	cross talk\\
\indent	impression show\\
\indent	autotune remix\\
\indent	meme\\
\indent	funny child\\
\indent	funny animal\\
\indent	funny event\\
\textbf{5. beauty}\\
\indent    beauty influencer\\
\indent	female model\\
\indent	layman beauty influencer\\
\textbf{6. handsome guy}\\
\indent	layman handsome influencer\\
\indent	male model\\
\indent	handsome influencer\\
\indent	young hunk\\
\indent	silver fox\\
\textbf{7. science \& technology}\\
\indent    science experiment\\
\indent	science anecdote\\
\indent	digital gadget\\
\indent	automation\\
\indent	scientific figure\\
\indent	cutting edge of science \& technology\\
\indent	aerospace\\
\indent	mechanical\\
\indent	blockchain\\
\textbf{8. sports}\\
\indent    fitness \& diet\\
\indent	basketball\\
\indent	billiard\\
\indent	yoga\\
\indent	football\\
\indent	water sports\\
\indent	ping-pong\\
\indent	run\\
\indent	volleyball\\
\indent	badminton\\
\indent	tennis\\
\indent	boxing\\
\indent	kong-fu\\
\indent	car racing\\
\indent	swim\\
\indent	cycling\\
\indent	extreme sports\\
\indent	chess \& card game\\
\indent	sports news\\
\indent	sports event\\
\indent	sports star\\
\indent	children's sport\\
\textbf{9. anime}\\
\indent    cosplay\\
\indent	anime cut\\
\indent	anime peripheral derivatives\\
\indent	manga\\
\indent	children's manga\\
\indent	One Piece\\
\indent	Naruto\\
\indent	Detective Conan\\
\textbf{10. game}\\
\indent    League of Legends\\
\indent	PUBG\\
\indent	Arena of Valor\\
\indent	Speed Drifters\\
\indent	Hearthstone\\
\indent	CSGO\\
\indent	Dota2\\
\indent	Overwatch\\
\textbf{11. vehicle}\\
 \indent   motorcycle\\
\indent	driving skill\\
\indent	vehicle maintenance\\
\indent	auto show\\
\indent	excavator\\
\indent	car tuning\\
\indent roadster\\
\indent	luxury car\\
\indent	traffic accident\\
\indent	driving test\\
\textbf{12. parenting}\\
\indent    pregnant mother\\
\indent	child education\\
\indent	child care\\
\indent	child talent\\
\indent	cute baby\\
\indent	parent-child interaction\\
\textbf{13. music}\\
\indent    singing\\
\indent	live music\\
\indent	music video\\
\indent	instrument playing\\
\indent	music teaching\\
\indent	music information\\
\indent	music review\\
\textbf{14. fashion}\\
\indent    hairdressing\\
\indent	nail art\\
\indent	skin care\\
\indent	outfit\\
\indent	make up\\
\indent	medical cosmetology\\
\indent	street shot\\
\indent	imitation makeup\\
\indent	men's fashion\\
\indent	fashion information\\
\textbf{15. society}\\
\indent    public welfare\\
\indent	natural disaster\\
\indent	anecdote\\
\indent	legal system\\
\indent	diplomacy\\
\indent	anti-corruption\\
\indent	regional affair\\
\indent	domestic affair\\
\indent	international news\\
\textbf{16. pet}\\
\indent    pet dog\\
\indent	pet cat\\
\indent	pet bird\\
\indent	pet reptile\\
\textbf{17. tourism}\\
\indent    customs\\
\indent	tourist attraction\\
\indent	tourist guide\\
\textbf{18. nature}\\
\indent    wild animal\\
\indent	wild plant\\
\textbf{19. daily life}\\
\indent    furniture\\
\indent	house decoration\\
\indent	good stuff\\
\indent	photo editing\\
\indent	ASMR\\
\indent	handwork\\
\indent	workplace\\
\indent	wedding ceremony\\
\indent	daily life tip\\
\indent	fun quiz\\
\textbf{20. finance}\\
  \indent  stock market\\
\indent	real estate\\
\indent	financial management\\
\indent	financial information\\
\indent	entrepreneurship\\
\indent	financial figure \\
\textbf{21. health}	\\
\indent    regimen\\
\indent	traditional medicine\\
\indent	medical science\\
\indent	sexual health\\
\textbf{22. military}\\
\indent    domestic military\\
\indent    intelligence\\
\indent	global military\\ 
\indent intelligence\\
\indent	military figure\\
\indent	armed special police\\
\indent	weaponry equipment\\
\indent	military exercise\\
\indent	war history\\
\textbf{23. history}\\
\indent    historical figure\\
\indent	world history\\
\indent	domestic history\\
\indent	archaeology \\
\textbf{24. education}\\
\indent    primary and secondary school\\
\indent	college and university \\
\indent	adult education \\
\indent	vocational examination \\
\indent	language teaching\\
\textbf{25. novelty seeking}\\
\indent    over-fancy mind experiment\\
\indent	oddity\\
\textbf{26. delicacy}\\
 \indent   mukbang\\
\indent	cooking tutorial\\
\indent	food review\\
\indent	restaurant review \\
\indent	snack\\
\indent	beverage\\
\textbf{27. culture and art}\\
 \indent   calligraphy\\
\indent    painting\\
\indent	acrobatics\\
\indent	magic\\
\indent	Go chess\\
\indent	reading\\
\indent	antique collection \\
\indent	opera\\
\indent	folklore\\
\indent	live theatre\\
\indent	sculpture\\
\indent	building\\
\indent	region\\
\textbf{28. horticulture}\\
\indent    flower arrangement \\
\indent	planting \\
\indent	plant science \\
\textbf{29. industry}\\
\indent    construction\\
\indent	super project\\
\indent	manufacture\\
\textbf{30. dance}\\
 \indent   K-pop dance\\
\indent	street dance\\
\indent	square dance\\
\indent	gesture dance\\
\indent	pole dance\\
\indent	ACG dance\\
\indent	folk dance\\
\indent	dance teaching\\
\textbf{31. affection}\\
 \indent   love\\
\indent	marriage\\
\indent	family\\
\indent	sociality\\
\indent	friendship\\
\textbf{32. countryside}\\
\indent    cultivation\\ 
\indent	hunting\\
\indent	fishing\\
\indent	farm work\\
\indent	grange\\
\indent	rural performance\\
\indent	rural custom\\
\indent	rural scenery\\
}
\newpage

{\small
\bibliographystyle{ieee_fullname}
\bibliography{egbib}

@String(CVPR  = {CVPR})

@String(ICCV  = {ICCV})

@String(ECCV  = {ECCV})

@String(ICASSP=	{ICASSP})

@String(AAAI = {AAAI})

@String(CVPR= {IEEE Conf. Comput. Vis. Pattern Recog.})

@String(ICCV= {Int. Conf. Comput. Vis.})

@String(ECCV= {Eur. Conf. Comput. Vis.})

@article{kay2017kinetics,
  title={The kinetics human action video dataset},
  author={Kay, Will and Carreira, Joao and Simonyan, Karen and Zhang, Brian and Hillier, Chloe and Vijayanarasimhan, Sudheendra and Viola, Fabio and Green, Tim and Back, Trevor and Natsev, Paul and others},
  journal={arXiv preprint arXiv:1705.06950},
  year={2017}
}

@article{monfort2019moments,
  title={Moments in time dataset: one million videos for event understanding},
  author={Monfort, Mathew and Andonian, Alex and Zhou, Bolei and Ramakrishnan, Kandan and Bargal, Sarah Adel and Yan, Tom and Brown, Lisa and Fan, Quanfu and Gutfreund, Dan and Vondrick, Carl and others},
  journal={IEEE transactions on pattern analysis and machine intelligence},
  volume={42},
  number={2},
  pages={502--508},
  year={2019},
  publisher={IEEE}
}

@article{Heilbron2015ActivityNetAL,
  title={ActivityNet: A large-scale video benchmark for human activity understanding},
  author={Fabian Caba Heilbron and Victor Escorcia and Bernard Ghanem and Juan Carlos Niebles},
  journal={2015 IEEE Conference on Computer Vision and Pattern Recognition (CVPR)},
  year={2015},
  pages={961-970}
}

@article{monfort2019multi,
  title={Multi-moments in time: Learning and interpreting models for multi-action video understanding},
  author={Monfort, Mathew and Ramakrishnan, Kandan and Andonian, Alex and McNamara, Barry A and Lascelles, Alex and Pan, Bowen and Fan, Quanfu and Gutfreund, Dan and Feris, Rogerio and Oliva, Aude},
  journal={arXiv preprint arXiv:1911.00232},
  year={2019}
}

@inproceedings{goyal2017something,
  title={The" Something Something" Video Database for Learning and Evaluating Visual Common Sense.},
  author={Goyal, Raghav and Kahou, Samira Ebrahimi and Michalski, Vincent and Materzynska, Joanna and Westphal, Susanne and Kim, Heuna and Haenel, Valentin and Fruend, Ingo and Yianilos, Peter and Mueller-Freitag, Moritz and others},
  booktitle={ICCV},
  volume={1},
  pages={5},
  year={2017}
}

@inproceedings{materzynska2019jester,
  title={The jester dataset: A large-scale video dataset of human gestures},
  author={Materzynska, Joanna and Berger, Guillaume and Bax, Ingo and Memisevic, Roland},
  booktitle={Proceedings of the IEEE International Conference on Computer Vision Workshops},
  pages={0--0},
  year={2019}
}

@article{soomro2012ucf101,
  title={UCF101: A dataset of 101 human actions classes from videos in the wild},
  author={Soomro, Khurram and Zamir, Amir Roshan and Shah, Mubarak},
  journal={arXiv preprint arXiv:1212.0402},
  year={2012}
}

@inproceedings{kuehne2011hmdb,
  title={HMDB: a large video database for human motion recognition},
  author={Kuehne, Hildegard and Jhuang, Hueihan and Garrote, Est{\'\i}baliz and Poggio, Tomaso and Serre, Thomas},
  booktitle={2011 International Conference on Computer Vision},
  pages={2556--2563},
  year={2011},
  organization={IEEE}
}

@article{abu2016youtube,
  title={Youtube-8m: A large-scale video classification benchmark},
  author={Abu-El-Haija, Sami and Kothari, Nisarg and Lee, Joonseok and Natsev, Paul and Toderici, George and Varadarajan, Balakrishnan and Vijayanarasimhan, Sudheendra},
  journal={arXiv preprint arXiv:1609.08675},
  year={2016}
}

@article{diba2019holistic,
  title={Holistic large scale video understanding},
  author={Diba, Ali and Fayyaz, Mohsen and Sharma, Vivek and Paluri, Manohar and Gall, Jurgen and Stiefelhagen, Rainer and Van Gool, Luc},
  journal={arXiv preprint arXiv:1904.11451},
  year={2019}
}

@inproceedings{karpathy2014large,
  title={Large-scale video classification with convolutional neural networks},
  author={Karpathy, Andrej and Toderici, George and Shetty, Sanketh and Leung, Thomas and Sukthankar, Rahul and Fei-Fei, Li},
  booktitle={Proceedings of the IEEE conference on Computer Vision and Pattern Recognition},
  pages={1725--1732},
  year={2014}
}

@inproceedings{ray2018scenes,
  title={Scenes-objects-actions: A multi-task, multi-label video dataset},
  author={Ray, Jamie and Wang, Heng and Tran, Du and Wang, Yufei and Feiszli, Matt and Torresani, Lorenzo and Paluri, Manohar},
  booktitle={Proceedings of the European Conference on Computer Vision (ECCV)},
  pages={635--651},
  year={2018}
}

@inproceedings{arandjelovic2017look,
  title={Look, listen and learn},
  author={Arandjelovic, Relja and Zisserman, Andrew},
  booktitle={Proceedings of the IEEE International Conference on Computer Vision},
  pages={609--617},
  year={2017}
}

@inproceedings{miech2019howto100m,
  title={Howto100m: Learning a text-video embedding by watching hundred million narrated video clips},
  author={Miech, Antoine and Zhukov, Dimitri and Alayrac, Jean-Baptiste and Tapaswi, Makarand and Laptev, Ivan and Sivic, Josef},
  booktitle={Proceedings of the IEEE international conference on computer vision},
  pages={2630--2640},
  year={2019}
}

@inproceedings{simonyan2014two,
  title={Two-stream convolutional networks for action recognition in videos},
  author={Simonyan, Karen and Zisserman, Andrew},
  booktitle={Advances in neural information processing systems},
  pages={568--576},
  year={2014}
}

@article{devlin2018bert,
  title={Bert: Pre-training of deep bidirectional transformers for language understanding},
  author={Devlin, Jacob and Chang, Ming-Wei and Lee, Kenton and Toutanova, Kristina},
  journal={arXiv preprint arXiv:1810.04805},
  year={2018}
}

@article{xiao2020audiovisual,
  title={Audiovisual SlowFast Networks for Video Recognition},
  author={Xiao, Fanyi and Lee, Yong Jae and Grauman, Kristen and Malik, Jitendra and Feichtenhofer, Christoph},
  journal={arXiv preprint arXiv:2001.08740},
  year={2020}
}

@inproceedings{carreira2017quo,
  title={Quo vadis, action recognition? a new model and the kinetics dataset},
  author={Carreira, Joao and Zisserman, Andrew},
  booktitle={proceedings of the IEEE Conference on Computer Vision and Pattern Recognition},
  pages={6299--6308},
  year={2017}
}

@inproceedings{feichtenhofer2019slowfast,
  title={Slowfast networks for video recognition},
  author={Feichtenhofer, Christoph and Fan, Haoqi and Malik, Jitendra and He, Kaiming},
  booktitle={Proceedings of the IEEE international conference on computer vision},
  pages={6202--6211},
  year={2019}
}

@inproceedings{lin2019tsm,
  title={Tsm: Temporal shift module for efficient video understanding},
  author={Lin, Ji and Gan, Chuang and Han, Song},
  booktitle={Proceedings of the IEEE International Conference on Computer Vision},
  pages={7083--7093},
  year={2019}
}

@inproceedings{wang2016temporal,
  title={Temporal segment networks: Towards good practices for deep action recognition},
  author={Wang, Limin and Xiong, Yuanjun and Wang, Zhe and Qiao, Yu and Lin, Dahua and Tang, Xiaoou and Van Gool, Luc},
  booktitle={European conference on computer vision},
  pages={20--36},
  year={2016},
  organization={Springer}
}

@inproceedings{liu2017towards,
  title={Towards micro-video understanding by joint sequential-sparse modeling},
  author={Liu, Meng and Nie, Liqiang and Wang, Meng and Chen, Baoquan},
  booktitle={Proceedings of the 25th ACM international conference on Multimedia},
  pages={970--978},
  year={2017}
}

@inproceedings{nie2017enhancing,
  title={Enhancing micro-video understanding by harnessing external sounds},
  author={Nie, Liqiang and Wang, Xiang and Zhang, Jianglong and He, Xiangnan and Zhang, Hanwang and Hong, Richang and Tian, Qi},
  booktitle={Proceedings of the 25th ACM international conference on Multimedia},
  pages={1192--1200},
  year={2017}
}

@article{wei2019neural,
  title={Neural multimodal cooperative learning toward micro-video understanding},
  author={Wei, Yinwei and Wang, Xiang and Guan, Weili and Nie, Liqiang and Lin, Zhouchen and Chen, Baoquan},
  journal={IEEE Transactions on Image Processing},
  volume={29},
  pages={1--14},
  year={2019},
  publisher={IEEE}
}

@article{nie2019multimodal,
  title={Multimodal learning toward micro-video understanding},
  author={Nie, Liqiang and Liu, Meng and Song, Xuemeng},
  journal={Synthesis Lectures on Image, Video, and Multimedia Processing},
  volume={9},
  number={4},
  pages={1--186},
  year={2019},
  publisher={Morgan \& Claypool Publishers}
}

@article{zhang2020low,
  title={Low-Rank Regularized Multimodal Representation for Micro-Video Event Detection},
  author={Zhang, Jing and Wu, Yuting and Liu, Jinghui and Jing, Peiguang and Su, Yuting},
  journal={IEEE Access},
  volume={8},
  pages={87266--87274},
  year={2020},
  publisher={IEEE}
}

@misc{easyOCR,
  title = {easyOCR},
  howpublished = {\url{https://github.com/JaidedAI/EasyOCR}},
}

@misc{textblob,
  title = {textblob},
  howpublished = {\url{https://github.com/sloria/TextBlob}},
}

@misc{ffmpeg,
  title = {FFMPeg},
  howpublished = {\url{www.ffmpeg.com}},
}

@misc{tiktok_official,
  title = {TikTok},
  howpublished = {\url{https://www.tiktok.com/}},
}

@misc{kwai_official,
  title = {Kwai},
  howpublished = {\url{https://www.kwai.com/}},
}

@misc{reels_official,
  title = {Reels},
  howpublished = {\url{https://about.instagram.com/blog/announcements/introducing-instagram-reels-announcement}},
}

@misc{fan2020pyslowfast,
  author = {Haoqi Fan and Yanghao Li and Bo Xiong and Wan-Yen Lo and
                  Christoph Feichtenhofer},
  title = {PySlowFast},
  howpublished = {\url{https://github.com/facebookresearch/slowfast}},
  year = {2020}
}

@misc{jiang2014thumos,
  title={THUMOS challenge: Action recognition with a large number of classes},
  author={Jiang, Yu-Gang and Liu, Jingen and Zamir, A Roshan and Toderici, George and Laptev, Ivan and Shah, Mubarak and Sukthankar, Rahul},
  year={2014}
}

@inproceedings{gu2018ava,
  title={Ava: A video dataset of spatio-temporally localized atomic visual actions},
  author={Gu, Chunhui and Sun, Chen and Ross, David A and Vondrick, Carl and Pantofaru, Caroline and Li, Yeqing and Vijayanarasimhan, Sudheendra and Toderici, George and Ricco, Susanna and Sukthankar, Rahul and others},
  booktitle={Proceedings of the IEEE Conference on Computer Vision and Pattern Recognition},
  pages={6047--6056},
  year={2018}
}

@misc{youtube_revenue,
  title = {Youtube revenue analysis.},
  howpublished = {\url{https://www.businessofapps.com/data/youtube-statistics/}},
}

@misc{tiktok,
title = {Tiktok statistics.}, 
howpublished =
{\url{https://www.oberlo.ca/blog/tiktok-statistics}}
}

@misc{mmaction2,
title = {mmaction2.}, 
howpublished =
{\url{https://github.com/open-mmlab/mmaction2/}}
}

@inproceedings{shao2020finegym,
  title={Finegym: A hierarchical video dataset for fine-grained action understanding},
  author={Shao, Dian and Zhao, Yue and Dai, Bo and Lin, Dahua},
  booktitle={Proceedings of the IEEE/CVF Conference on Computer Vision and Pattern Recognition},
  pages={2616--2625},
  year={2020}
}

@inproceedings{shahroudy2016ntu,
  title={Ntu rgb+ d: A large scale dataset for 3d human activity analysis},
  author={Shahroudy, Amir and Liu, Jun and Ng, Tian-Tsong and Wang, Gang},
  booktitle={Proceedings of the IEEE conference on computer vision and pattern recognition},
  pages={1010--1019},
  year={2016}
}

@inproceedings{oh2011large,
  title={A large-scale benchmark dataset for event recognition in surveillance video},
  author={Oh, Sangmin and Hoogs, Anthony and Perera, Amitha and Cuntoor, Naresh and Chen, Chia-Chih and Lee, Jong Taek and Mukherjee, Saurajit and Aggarwal, JK and Lee, Hyungtae and Davis, Larry and others},
  booktitle={CVPR 2011},
  pages={3153--3160},
  year={2011},
  organization={IEEE}
}

@article{liu2017pku,
  title={Pku-mmd: A large scale benchmark for continuous multi-modal human action understanding},
  author={Liu, Chunhui and Hu, Yueyu and Li, Yanghao and Song, Sijie and Liu, Jiaying},
  journal={arXiv preprint arXiv:1703.07475},
  year={2017}
}

@inproceedings{xu2018large,
  title={Large-scale weakly supervised audio classification using gated convolutional neural network},
  author={Xu, Yong and Kong, Qiuqiang and Wang, Wenwu and Plumbley, Mark D},
  booktitle={2018 IEEE international conference on acoustics, speech and signal processing (ICASSP)},
  pages={121--125},
  year={2018},
  organization={IEEE}
}

@article{shukla2020learning,
  title={Learning Speech Representations from Raw Audio by Joint Audiovisual Self-Supervision},
  author={Shukla, Abhinav and Petridis, Stavros and Pantic, Maja},
  journal={arXiv preprint arXiv:2007.04134},
  year={2020}
}

@inproceedings{tran2018closer,
  title={A closer look at spatiotemporal convolutions for action recognition},
  author={Tran, Du and Wang, Heng and Torresani, Lorenzo and Ray, Jamie and LeCun, Yann and Paluri, Manohar},
  booktitle={Proceedings of the IEEE conference on Computer Vision and Pattern Recognition},
  pages={6450--6459},
  year={2018}
}

@inproceedings{ibrahim2016hierarchical,
  title={A hierarchical deep temporal model for group activity recognition},
  author={Ibrahim, Mostafa S and Muralidharan, Srikanth and Deng, Zhiwei and Vahdat, Arash and Mori, Greg},
  booktitle={Proceedings of the IEEE Conference on Computer Vision and Pattern Recognition},
  pages={1971--1980},
  year={2016}
}

@inproceedings{zhu2013videotopic,
  title={Videotopic: Content-based video recommendation using a topic model},
  author={Zhu, Qiusha and Shyu, Mei-Ling and Wang, Haohong},
  booktitle={2013 IEEE International Symposium on Multimedia},
  pages={219--222},
  year={2013},
  organization={IEEE}
}

@article{deng2015twitter,
  title={Twitter is faster: Personalized time-aware video recommendation from twitter to youtube},
  author={Deng, Zhengyu and Yan, Ming and Sang, Jitao and Xu, Changsheng},
  journal={ACM Transactions on Multimedia Computing, Communications, and Applications (TOMM)},
  volume={11},
  number={2},
  pages={1--23},
  year={2015},
  publisher={ACM New York, NY, USA}
}

@inproceedings{davidson2010youtube,
  title={The YouTube video recommendation system},
  author={Davidson, James and Liebald, Benjamin and Liu, Junning and Nandy, Palash and Van Vleet, Taylor and Gargi, Ullas and Gupta, Sujoy and He, Yu and Lambert, Mike and Livingston, Blake and others},
  booktitle={Proceedings of the fourth ACM conference on Recommender systems},
  pages={293--296},
  year={2010}
}

@article{ji2019query,
  title={Query-aware sparse coding for web multi-video summarization},
  author={Ji, Zhong and Ma, Yaru and Pang, Yanwei and Li, Xuelong},
  journal={Information Sciences},
  volume={478},
  pages={152--166},
  year={2019},
  publisher={Elsevier}
}

@article{maaten2008visualizing,
  title={Visualizing data using t-SNE},
  author={Maaten, Laurens van der and Hinton, Geoffrey},
  journal={Journal of machine learning research},
  volume={9},
  number={Nov},
  pages={2579--2605},
  year={2008}
}

@inproceedings{He_2016_CVPR,
author = {He, Kaiming and Zhang, Xiangyu and Ren, Shaoqing and Sun, Jian},
title = {Deep Residual Learning for Image Recognition},
booktitle = {Proceedings of the IEEE Conference on Computer Vision and Pattern Recognition (CVPR)},
month = {June},
year = {2016}
}

@inproceedings{lin2014visual,
  title={Visual semantic search: Retrieving videos via complex textual queries},
  author={Lin, Dahua and Fidler, Sanja and Kong, Chen and Urtasun, Raquel},
  booktitle={Proceedings of the IEEE conference on computer vision and pattern recognition},
  pages={2657--2664},
  year={2014}
}

@article{kiros2014unifying,
  title={Unifying visual-semantic embeddings with multimodal neural language models},
  author={Kiros, Ryan and Salakhutdinov, Ruslan and Zemel, Richard S},
  journal={arXiv preprint arXiv:1411.2539},
  year={2014}
}

@article{torabi2016learning,
  title={Learning language-visual embedding for movie understanding with natural-language},
  author={Torabi, Atousa and Tandon, Niket and Sigal, Leonid},
  journal={arXiv preprint arXiv:1609.08124},
  year={2016}
}

@inproceedings{otani2016learning,
  title={Learning joint representations of videos and sentences with web image search},
  author={Otani, Mayu and Nakashima, Yuta and Rahtu, Esa and Heikkil{\"a}, Janne and Yokoya, Naokazu},
  booktitle={European Conference on Computer Vision},
  pages={651--667},
  year={2016},
  organization={Springer}
}

@inproceedings{yu2017end,
  title={End-to-end concept word detection for video captioning, retrieval, and question answering},
  author={Yu, Youngjae and Ko, Hyungjin and Choi, Jongwook and Kim, Gunhee},
  booktitle={Proceedings of the IEEE Conference on Computer Vision and Pattern Recognition},
  pages={3165--3173},
  year={2017}
}

@article{miech2018learning,
  title={Learning a text-video embedding from incomplete and heterogeneous data},
  author={Miech, Antoine and Laptev, Ivan and Sivic, Josef},
  journal={arXiv preprint arXiv:1804.02516},
  year={2018}
}

@inproceedings{yu2018joint,
  title={A joint sequence fusion model for video question answering and retrieval},
  author={Yu, Youngjae and Kim, Jongseok and Kim, Gunhee},
  booktitle={Proceedings of the European Conference on Computer Vision (ECCV)},
  pages={471--487},
  year={2018}
}

@article{lei2021less,
  title={Less is More: ClipBERT for Video-and-Language Learning via Sparse Sampling},
  author={Lei, Jie and Li, Linjie and Zhou, Luowei and Gan, Zhe and Berg, Tamara L and Bansal, Mohit and Liu, Jingjing},
  journal={arXiv preprint arXiv:2102.06183},
  year={2021}
}

@inproceedings{xu2016msr,
  title={Msr-vtt: A large video description dataset for bridging video and language},
  author={Xu, Jun and Mei, Tao and Yao, Ting and Rui, Yong},
  booktitle={Proceedings of the IEEE conference on computer vision and pattern recognition},
  pages={5288--5296},
  year={2016}
}

@article{rohrbach2017movie,
  title={Movie description},
  author={Rohrbach, Anna and Torabi, Atousa and Rohrbach, Marcus and Tandon, Niket and Pal, Christopher and Larochelle, Hugo and Courville, Aaron and Schiele, Bernt},
  journal={International Journal of Computer Vision},
  volume={123},
  number={1},
  pages={94--120},
  year={2017},
  publisher={Springer}
}

@inproceedings{zhou2018towards,
  title={Towards automatic learning of procedures from web instructional videos},
  author={Zhou, Luowei and Xu, Chenliang and Corso, Jason},
  booktitle={Proceedings of the AAAI Conference on Artificial Intelligence},
  volume={32},
  number={1},
  year={2018}
}

@inproceedings{damen2018scaling,
  title={Scaling egocentric vision: The epic-kitchens dataset},
  author={Damen, Dima and Doughty, Hazel and Farinella, Giovanni Maria and Fidler, Sanja and Furnari, Antonino and Kazakos, Evangelos and Moltisanti, Davide and Munro, Jonathan and Perrett, Toby and Price, Will and others},
  booktitle={Proceedings of the European Conference on Computer Vision (ECCV)},
  pages={720--736},
  year={2018}
}

@inproceedings{krishna2017dense,
  title={Dense-captioning events in videos},
  author={Krishna, Ranjay and Hata, Kenji and Ren, Frederic and Fei-Fei, Li and Carlos Niebles, Juan},
  booktitle={Proceedings of the IEEE international conference on computer vision},
  pages={706--715},
  year={2017}
}

@inproceedings{anne2017localizing,
  title={Localizing moments in video with natural language},
  author={Anne Hendricks, Lisa and Wang, Oliver and Shechtman, Eli and Sivic, Josef and Darrell, Trevor and Russell, Bryan},
  booktitle={Proceedings of the IEEE international conference on computer vision},
  pages={5803--5812},
  year={2017}
}

@inproceedings{fernando2017self,
  title={Self-supervised video representation learning with odd-one-out networks},
  author={Fernando, Basura and Bilen, Hakan and Gavves, Efstratios and Gould, Stephen},
  booktitle={Proceedings of the IEEE conference on computer vision and pattern recognition},
  pages={3636--3645},
  year={2017}
}

@inproceedings{lee2017unsupervised,
  title={Unsupervised representation learning by sorting sequences},
  author={Lee, Hsin-Ying and Huang, Jia-Bin and Singh, Maneesh and Yang, Ming-Hsuan},
  booktitle={Proceedings of the IEEE International Conference on Computer Vision},
  pages={667--676},
  year={2017}
}

@inproceedings{han2019video,
  title={Video representation learning by dense predictive coding},
  author={Han, Tengda and Xie, Weidi and Zisserman, Andrew},
  booktitle={Proceedings of the IEEE/CVF International Conference on Computer Vision Workshops},
  pages={0--0},
  year={2019}
}

@inproceedings{wang2019self,
  title={Self-supervised spatio-temporal representation learning for videos by predicting motion and appearance statistics},
  author={Wang, Jiangliu and Jiao, Jianbo and Bao, Linchao and He, Shengfeng and Liu, Yunhui and Liu, Wei},
  booktitle={Proceedings of the IEEE/CVF Conference on Computer Vision and Pattern Recognition},
  pages={4006--4015},
  year={2019}
}

@article{sun2019learning,
  title={Learning video representations using contrastive bidirectional transformer},
  author={Sun, Chen and Baradel, Fabien and Murphy, Kevin and Schmid, Cordelia},
  journal={arXiv preprint arXiv:1906.05743},
  year={2019}
}

@inproceedings{miech2020end,
  title={End-to-end learning of visual representations from uncurated instructional videos},
  author={Miech, Antoine and Alayrac, Jean-Baptiste and Smaira, Lucas and Laptev, Ivan and Sivic, Josef and Zisserman, Andrew},
  booktitle={Proceedings of the IEEE/CVF Conference on Computer Vision and Pattern Recognition},
  pages={9879--9889},
  year={2020}
}

@article{radford2021learning,
  title={Learning transferable visual models from natural language supervision},
  author={Radford, Alec and Kim, Jong Wook and Hallacy, Chris and Ramesh, Aditya and Goh, Gabriel and Agarwal, Sandhini and Sastry, Girish and Askell, Amanda and Mishkin, Pamela and Clark, Jack and others},
  journal={arXiv preprint arXiv:2103.00020},
  year={2021}
}

@article{korbar2018cooperative,
  title={Cooperative learning of audio and video models from self-supervised synchronization},
  author={Korbar, Bruno and Tran, Du and Torresani, Lorenzo},
  journal={arXiv preprint arXiv:1807.00230},
  year={2018}
}

@article{li2020learning,
  title={Learning spatiotemporal features via video and text pair discrimination},
  author={Li, Tianhao and Wang, Limin},
  journal={arXiv preprint arXiv:2001.05691},
  year={2020}
}

@article{damen2020epic,
  title={The epic-kitchens dataset: Collection, challenges and baselines},
  author={Damen, Dima and Doughty, Hazel and Farinella, Giovanni Maria and Fidler, Sanja and Furnari, Antonino and Kazakos, Evangelos and Moltisanti, Davide and Munro, Jonathan and Perrett, Toby and Price, Will and others},
  journal={IEEE Transactions on Pattern Analysis and Machine Intelligence},
  volume={43},
  number={11},
  pages={4125--4141},
  year={2020},
  publisher={IEEE}
}

@article{carreira2018short,
  title={A short note about kinetics-600},
  author={Carreira, Joao and Noland, Eric and Banki-Horvath, Andras and Hillier, Chloe and Zisserman, Andrew},
  journal={arXiv preprint arXiv:1808.01340},
  year={2018}
}

@inproceedings{diba2020large,
  title={Large scale holistic video understanding},
  author={Diba, Ali and Fayyaz, Mohsen and Sharma, Vivek and Paluri, Manohar and Gall, J{\"u}rgen and Stiefelhagen, Rainer and Gool, Luc Van},
  booktitle={European Conference on Computer Vision},
  pages={593--610},
  year={2020},
  organization={Springer}
}

@inproceedings{KarpathyCVPR14,
  title     = {Large-scale Video Classification with Convolutional Neural Networks},
  author    = {Andrej Karpathy and George Toderici and Sanketh Shetty and Thomas Leung and Rahul Sukthankar and Li Fei-Fei},
  year      = {2014},
  booktitle = {CVPR}
}

@inproceedings{rohrbach2015dataset,
  title={A dataset for movie description},
  author={Rohrbach, Anna and Rohrbach, Marcus and Tandon, Niket and Schiele, Bernt},
  booktitle={Proceedings of the IEEE conference on computer vision and pattern recognition},
  pages={3202--3212},
  year={2015}
}
}

\end{document}